\documentclass[10pt,letterpaper]{article}

\usepackage[pagenumbers]{cvpr} %

\usepackage{listings}
\usepackage{xcolor}

\definecolor{cvprblue}{rgb}{0.21,0.49,0.74}
\usepackage[pagebackref,breaklinks,colorlinks,allcolors=cvprblue]{hyperref}
\usepackage{subcaption}
\def\paperID{*****} %
\def\confName{CVPR}
\def\confYear{2025}

\title{UniRL-Zero: Reinforcement Learning on Unified Models \\ with Joint Language Model and Diffusion Model Experts}

\author{
Fu-Yun Wang\textsuperscript{*} \\
{\tt\small fywang0126@gmail.com} 
\and
Han Zhang\textsuperscript{*} \\
{\tt\small hanzhang.ai@gmail.com } 
\and 
Michaël Gharbi \\
{\tt\small michael.yanis.gharbi@gmail.com}
\and 
Hongsheng Li \\
{\tt\small hsli@ee.cuhk.edu.hk}
\and 
Taesung Park \\
{\tt\small taesung89@gmail.com}
}

\begin{document}
\maketitle

\begin{abstract}
We present UniRL-Zero, a unified reinforcement learning (RL) framework that boosts, multimodal language model understanding and reasoning, diffusion model multimedia generation, and their beneficial interaction capabilities within a unified model. Our work defines six scenarios for unified model reinforcement learning, providing systematic baselines for reinforcement learning of unified understanding and generation model. Our code is available at \url{https://github.com/G-U-N/UniRL}.
\end{abstract}
    
\begin{center}
    \includegraphics[width=0.99\linewidth]{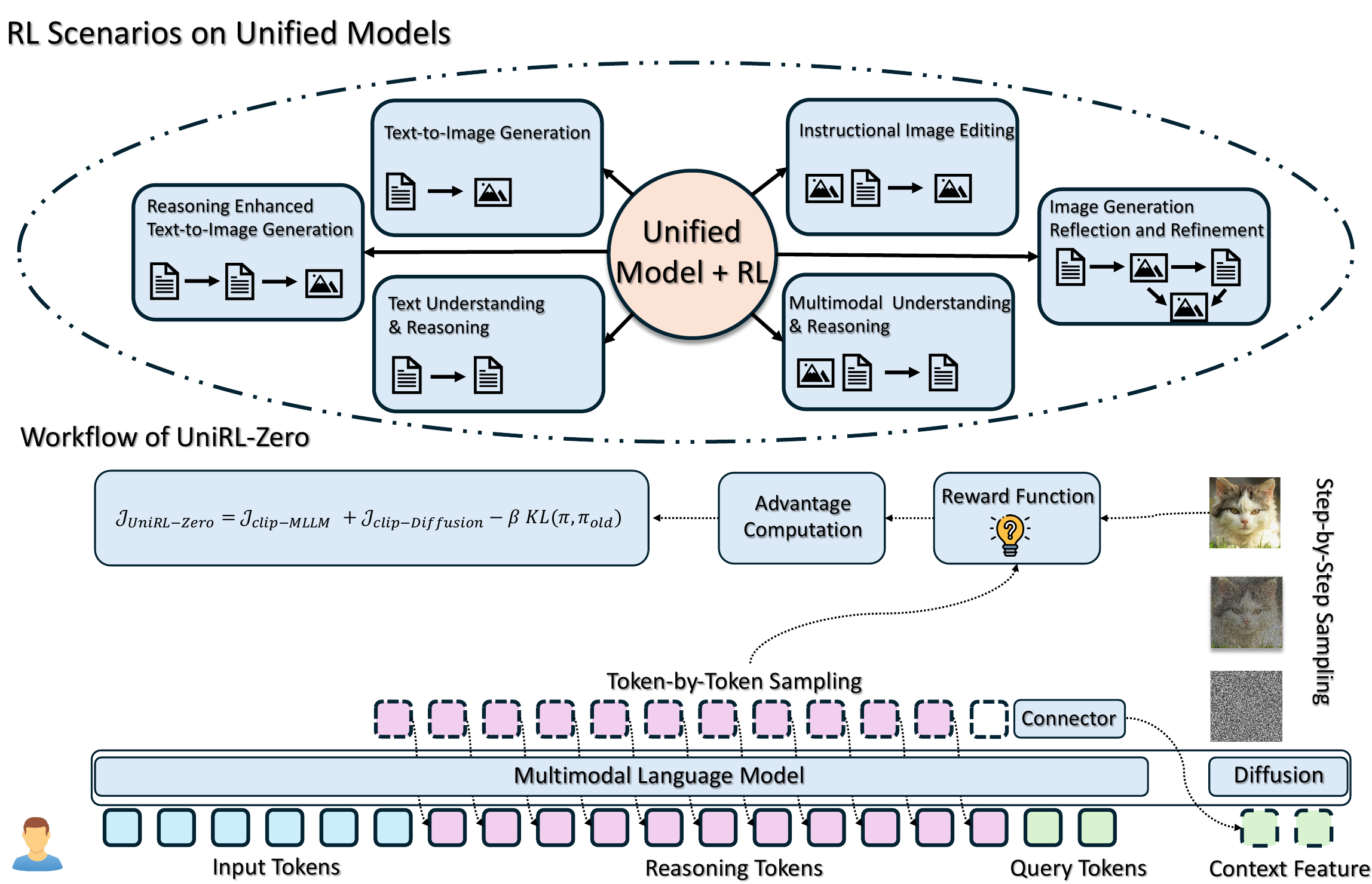}
    \captionof{figure}{Overview of the UniRL-Zero framework. }
    \label{fig:placeholder}
\end{center}

\tableofcontents
\thispagestyle{empty} %
\clearpage %
\setcounter{page}{1} %
\section{Introduction}
\label{sec:intro}

The rapid progress of generative AI in recent years has been largely driven by two foundational model families.

\textbf{Language Models (LMs)}—such as GPT\cite{brown2020language} and Gemini~\cite{team2023gemini,team2024gemini}—have significantly advanced natural language understanding and multimodal reasoning, enabling rich contextual interpretation and complex tasks.

\textbf{Diffusion Models (DMs)}—including VEO~\cite{wiedemer2025video}, SORA~\cite{brooks2024video}, Nano Banana~\cite{google_nano_banana_2025}, and GPT-4o image~\cite{hurst2024gpt}—have transformed multimedia generation, achieving high fidelity and realism in images, videos, and other media.

Reinforcement Learning (RL)~\cite{sutton1998reinforcement} has played a pivotal role in this evolution. Since the release of ChatGPT~\cite{openai_chatgpt} with its Reinforcement Learning from Human Feedback (RLHF) approach~\cite{ouyang2022training}, RL has become essential for aligning generative models with human preferences. For LMs, large-scale RL techniques, such as GRPO in DeepSeekMath~\cite{shao2024deepseekmath}, have improved reasoning and comprehension. For DMs, methods like DDPO~\cite{black2023training}, Flow-GRPO~\cite{liu2025flow}, and Dance-GRPO~\cite{xue2025dancegrpo} have enhanced preference alignment and compositional quality in generated content.

As generative AI matures, there is increasing interest in unified models that integrate the reasoning capabilities of LMs with the generative power of DMs. Recent works, such as Bagel~\cite{deng2025emerging} and Next-Step~\cite{team2025nextstep}, demonstrate the potential of such hybrid systems, where LMs handle structured reasoning and DMs deliver high-quality content generation.

Despite these advances, reinforcement learning for unified models remains underexplored. Existing strategies do not adequately address the joint optimization of LMs and DMs within a single framework, limiting the ability to fully leverage their complementary strengths.

To address this gap, we introduce \textbf{UniRL-Zero}, a unified reinforcement learning framework designed to enhance  language model understanding and reasoning, diffusion model multimedia generation, and their synergistic interaction. We define six core scenarios for reinforcement learning in unified models and provide systematic baselines to guide future research.

\section{How is RL formulated in LMs and DMs?}
\label{sec:bg}

\textbf{LMs (Discrete token-level RL).}
For LMs, RL is formulated at the token level. The LM serves as the agent, with the state defined by the sequence of tokens generated so far. Each action corresponds to selecting the next token from the vocabulary, according to the model’s policy distribution. The objective is to optimize the LM’s policy to produce sequences that maximize expected reward, often using policy gradient methods like PPO~\cite{schulman2017proximal} or GRPO.

\textbf{DMs (Continuous denoising-step RL).}
In DMs, RL is applied at the noise timestep level. DMs generate images by reversing a stochastic differential equation (SDE)~\cite{ho2020denoising,song2020score,karras2022elucidating,wang2024rectified,liu2022rectified}. The denoising network acts as the agent, where the state is the noisy image at a given timestep. The policy governs the denoising trajectory, transforming pure noise into a clean image. Rewards are generally assigned at the sequence level—evaluating the final output’s aesthetics or prompt alignment—rather than at each denoising step.  Unlike LMs, DMs adopt a continuous action space, as the model outputs vectors rather than discrete tokens.

\section{How is RL formulated in unified models with joint LM and DM experts?}

Unified models integrate LMs for reasoning and DMs for high-fidelity multimedia generation. This joint framework enables end-to-end optimization where both modules act as cooperative experts. RL in this setting extends beyond isolated token-level or timestep-level optimization, encompassing inter-module interactions. The overall policy thus combines discrete actions from LMs (e.g., generating reasoning steps or captions) with continuous actions from DMs (e.g., denoising trajectories).

We define six core RL scenarios in unified models, each capturing a distinct integration of understanding and generation:

\textbf{Scenario 1: text understanding and reasoning (text $\rightarrow$ text).}
The LM alone handles token-level prediction for tasks such as question answering or summarization. RL directly optimizes the LM’s discrete policy for qualities like correctness, reasoning depth, and helpfulness.

\textbf{Scenario 2: multimodal reasoning (image + text $\rightarrow$ text).}
The LM integrates visual and textual features to produce text outputs, e.g., image captioning or multimodal analysis. RL focuses on accuracy and coherence in multimodal reasoning, with no DM involvement.

\textbf{Scenario 3: text-to-image generation (text $\rightarrow$ image).}
The text prompt is encoded by the LM into semantic features or query tokens, which condition the DM. The DM executes the denoising trajectory to synthesize images. RL rewards target alignment with the prompt and visual quality, while the LM functions only as a semantic encoder.

\textbf{Scenario 4: instructional image editing (text + image $\rightarrow$ image).}
The LM converts editing instructions into conditioning features. The DM, given the source image (and optionally a mask), performs the editing trajectory. RL evaluates the final image for both instruction compliance and preservation of original content, with the LM again acting as a semantic feature provider.

\textbf{Scenario 5: CoT-enhanced text-to-image (text $\rightarrow$ text (reasoning) $\rightarrow$ image).}
The LM first performs reasoning to produce structured or intermediate text, which is then passed as semantic input to the DM. The DM generates the final image, while RL jointly optimizes reasoning quality and visual alignment.

\textbf{Scenario 6: reflective image generation (text $\rightarrow$ image $\rightarrow$ text (reflection) $\rightarrow$ image).}
The LM and DM interact iteratively: the DM generates an image from text, the LM reflects on the result and produces feedback, and the DM refines the image accordingly. RL encourages improvements across cycles, rewarding better alignment and refinement with each iteration.

Scenarios 1–2 (text and multimodal reasoning) are relatively well-studied. This work focuses on Scenarios 3–6, where generative tasks require tight synergy between LMs and DMs, making effective RL design particularly critical.

\section{RL on unified models with joint LM and DM experts}

\subsection{Base unified model}
To explore the effectiveness of RL strategies on the scenarios described above, we need to train a base unified model that integrates joint LM and DM experts. We follow the idea of MetaQuery~\cite{pan2025transfermodalitiesmetaqueries} considering its simplicity and training efficiency. We provide the implementation details in Section~\ref{sec:basemodel}.

\subsection{RL on unified models with joint LM and DM experts}

To formalize the reinforcement learning (RL) in our unified framework, we consider a joint policy optimization problem that integrates the discrete token-level actions of the LM with the continuous denoising actions of the DM, operating on interleaved text-image data. The goal is to optimize the end-to-end policy to maximize expected rewards derived from generated textual outputs and corresponding visual content, ensuring coherent and high-quality interleaved text-image sequences.

Formally, let $\mathcal{Q}$ denote a query, comprising a textual input $q_{\text{text}}$ (\eg, a caption or instruction) and visual input $\mathbf{q}_{\text{image}}$ (\eg, a reference image for image editing). 

The RL process unfolds as follows:

\begin{enumerate}
\item \textbf{LM reasoning}: The LM, parameterized by $\theta_{\text{LM}}$, processes the query to generate a reasoning sequence. The output is a reasoned text sequence $y_{\text{reason}} = [a_1, a_2, \dots, a_T]$, which includes structured elements like chain-of-thought (CoT) (e.g., \texttt{<think>} tags) or answer markers (e.g., \texttt{<answer>} tags) to improve interpretability.

\item \textbf{Context extraction}: Following MetaQuery, we employ trainable meta-query tokens. These tokens attend to the LM's hidden states via cross-attention, extracting query-specific features. A bi-directional connector transformer further refines these features.

\item \textbf{DM sampling}: The extracted context features $\mathbf{f}$ condition the DM. Starting from pure noise $\mathbf{x}_1 \sim \mathcal{N}(\mathbf{0}, \mathbf{I})$, the DM predicts denoising steps via a reversal  stochastic differential equation (SDE) process. Specifically, 
   \begin{equation}
   \mathbf{x}_{t-d t} = \mathbf{x}_t \cdot \left(1 - \frac{D_t}{1 - t}\right) + \mathbf{v}_t \cdot (d t - D_t) + \sqrt{2 \cdot D_t \cdot \frac{t}{1 - t}} \cdot \boldsymbol{\epsilon}, \quad \boldsymbol \epsilon \sim \mathcal{N}(\mathbf{0}, \mathbf{I})\,
   \end{equation}
   where $D_t$ can be set as $D_t = \eta dt$ (See our intuitive proof in Supplementary Section~\ref{sec:sdeproof}).  The output is the denoising trajectory $y_{\text{denoise}} = [\mathbf{x}_1, \mathbf{x}_{1-dt}, \mathbf{x}_{1-2dt}, \dots, \mathbf{x}_0 ]$.

\item  \textbf{Generated image reflection}:  The generated image  $\mathbf{x}_0$ is fed back as an additional input to the LM. The LM processes this visual input alongside the original query and prior reasoning $y_{\text{reason}}$ to generate a reflection sequence $y_{\text{reflect}} = [a_{T+1}, a_{T+2}, \dots, a_{T+M}]$, analyzing potential issues and suggesting refinements. This trigger another cycle of context extraction and DM sampling, forming an iterative loop until a termination condition (e.g., reward threshold or max iterations) is met.

\end{enumerate}

\noindent \textbf{Policy optimization}: The unified policy $\pi_{\theta} = \{ \pi_{\theta_\text{LM}}, \pi_{\theta_\text{DM}}\}$ governs the composite trajectory $\tau = (\mathcal Q, \tau_{\text{LM}}, \tau_{\text{DM}}, \tau_{\text{LM}}', \tau_{\text{DM}}', \dots )$, where $\tau_{\text{LM}} = [a_0,  a_1, \dots, a_T]$ is the discrete token trajectory, and $\tau_{\text{DM}} = (\mathbf{x}_1,  \mathbf{x}_{1-dt},  \dots, \mathbf{x}_0)$ is the continuous denoising trajectory.   A reward $R(\tau)$ evaluates the complete trajectory, which might comprise multiple components for holistic alignment.

We employ Group Relative Policy Optimization (GRPO) for its simplicity and efficiency. For each query $\mathcal{Q}$, we sample a group of $G$ trajectories $\{\tau_i\}_{i=1}^G$ from the old policy $\pi_{\theta_{\text{old}}}$, compute rewards $\{R(\tau_i)\}$, and derive normalized advantages $\hat{A}_i = (R(\tau_i) - \overline{R}) / \sigma_R$, where $\overline{R}$ and $\sigma_R$ are the group mean and standard deviation. We assign the same advantage $\hat{A}_i$ to all actions in $\tau_i$.

To compute the loss per-action, we approximate the joint optimization by decomposing into module-specific surrogates, averaging over actions:
\begin{itemize}

\item \textbf{LM loss}: 
\begin{equation}
\mathcal{J}_{\text{clip-MLLM}} = -\mathbb{E}_{\tau} \left[ \frac{1}{T} \sum_{t=1}^T \min\left( r_t^{\text{LM}} \hat{A}(\tau), \text{clip}(r_t^{\text{LM}}, 1-\epsilon_{\text{LM}}, 1+\epsilon_{\text{LM}}) \hat{A}(\tau) \right) \right]
\end{equation}
, where $r_t^{\text{LM}} = \frac{\pi_{\theta_{\text{LM}}}}{\pi_{\theta_{\text{old, LM}}}}$ is the per-token probability ratio.

\item \textbf{DM loss}: \begin{equation}\mathcal{J}_{\text{clip-Diffusion}} = -\mathbb{E}_{\tau} \left[ \frac{1}{N} \sum_{k=1}^N \min\left( r_k^{\text{DM}} \hat{A}(\tau),\text{clip}(r_k^{\text{DM}}, 1-\epsilon_{\text{DM}}, 1+\epsilon_{\text{DM}}) \hat{A}(\tau) \right) \right]\end{equation}
, where $r_k^{\text{DM}} = \frac{\pi_{\theta_{\text{DM}}}}{\pi_{\theta_{\text{old, DM}}}}$ is the per-step density ratio.
\end{itemize}

A KL divergence term $\beta \mathbb{E} [D_{\text{KL}}(\pi_{\theta} || \pi_{\text{ref}})]$ is added to the total loss to prevent deviation from a reference policy $\pi_{\text{ref}}$. Gradients are computed end-to-end, updating $\theta_{\text{LM}}$, $\theta_{\text{conn}}$, and $\theta_{\text{DM}}$ jointly.

\section{Experiments}

\subsection{Performance of the base model}
Table~\ref{tab:image_generation} compares our base model against several established models on image generation tasks. The metrics evaluate overall performance on GenEval~\cite{ghosh2023geneval} including specific capabilities such as single object generation, counting, color accuracy, two-object composition, positional accuracy, and color attribution. Our model, trained at 1024$\times$1024 resolution, achieves an overall score of 0.69, outperforming models like PixArt-$\alpha$ (0.48), PixArt-$\Sigma$ (0.52), and even larger models like LUMINA-Next (0.46) and SDXL (0.55). 

\begin{table}[h]

\centering

\caption{Performance comparison of our base model against other models on text-to-image composition ability.}

\label{tab:image_generation}

\resizebox{0.95\textwidth}{!}{
\begin{tabular}{l c c c c c c c c c}

\toprule

\textbf{Model/Configuration} & \textbf{Params (B)} & \textbf{Resolution} & \textbf{Overall} & \textbf{Single Object} & \textbf{Counting} & \textbf{Colors} & \textbf{Two Object} & \textbf{Position} & \textbf{Color Attribution} \\

\midrule

\textbf{Ours} & \textbf{1.6} & \textbf{1024$\times$1024} & \textbf{0.69} & \textbf{1.00} & \textbf{0.62} & \textbf{0.87} & \textbf{0.85} & \textbf{0.32} & \textbf{0.58} \\

PixArt-$\alpha$~\cite{chen2023pixart} & 0.6 & 512$\times$512 & 0.48 & 0.98 & 0.50 & 0.44 & 0.80 & 0.08 & 0.07 \\

PixArt-$\Sigma$~\cite{chen2024pixart} & 0.6 & 512$\times$512 & 0.52 & 0.98 & 0.59 & 0.50 & 0.80 & 0.10 & 0.15 \\

Sana-0.6B~\cite{xie2024sana} & 0.6 & 1024$\times$1024 & 0.64 & 0.99 & 0.76 & 0.64 & 0.88 & 0.18 & 0.39 \\

Sana-1.6B & 1.6 & 1024$\times$1024 & 0.66 & 0.99 & 0.77 & 0.62 & 0.88 & 0.21 & 0.47 \\

LUMINA-Next~\cite{zhuo2024lumina} & 2.0 & 1024$\times$1024 & 0.46 & 0.92 & 0.46 & 0.48 & 0.70 & 0.09 & 0.13 \\

SDXL~\cite{podell2023sdxl} & 2.6 & 1024$\times$1024 & 0.55 & 0.98 & 0.74 & 0.39 & 0.85 & 0.15 & 0.23 \\

PlayGroundv2.5 & 2.6 & 1024$\times$1024 & 0.56 & 0.98 & 0.77 & 0.52 & 0.84 & 0.11 & 0.17 \\

Hunyuan-DiT~\cite{li2024hunyuan} & 1.5 & 1024$\times$1024 & 0.63 & 0.97 & 0.77 & 0.71 & 0.88 & 0.13 & 0.30 \\

DALLE3~\cite{betker2023improving} & - & 1024$\times$1024 & 0.67 & 0.96 & 0.87 & 0.47 & 0.83 & 0.43 & 0.45 \\

SD3-medium~\cite{esser2024scaling} & 2.0 & 1024$\times$1024 & 0.62 & 0.98 & 0.74 & 0.63 & 0.67 & 0.34 & 0.36 \\

FLUX-dev~\cite{batifol2025flux} & 12.0 & 1024$\times$1024 & 0.67 & 0.99 & 0.81 & 0.79 & 0.74 & 0.20 & 0.47 \\

FLUX-schnell~\cite{blackforestlabs2024flux} & 12.0 & 1024$\times$1024 & 0.71 & 0.99 & 0.92 & 0.73 & 0.78 & 0.28 & 0.54 \\

\bottomrule

\end{tabular}
}
\end{table}

Table~\ref{tab:multimodal_reasoning} evaluates the base model's multimodal reasoning capabilities across standard benchmarks like MME-P~\cite{yin2024survey}, MMB~\cite{liu2024mmbench}, SEED~\cite{li2023seed}, MMMU~\cite{yue2024mmmu}, and MM-Vet~\cite{yu2023mm}. Our model, built on Qwen2.5-VL 3B~\cite{bai2025qwen2}, achieves a strong MM-Vet score of 63.2, surpassing models like Janus-Pro-7B (50.0) and TokenFlow-XL (48.2). It also demonstrates competitive performance on MME-P (1574.3) and SEED (73.8), indicating robust text and multimodal reasoning abilities.

\begin{table}[h]

\centering

\caption{Performance comparison of our base model against other models on multimodal reasoning tasks.}

\label{tab:multimodal_reasoning}

\begin{tabular}{l l c c c c c}

\toprule

\textbf{Methods} & \textbf{Base (M)LLM} & \textbf{MME-P} & \textbf{MMB} & \textbf{SEED} & \textbf{MMMU} & \textbf{MM-Vet} \\

\midrule

Emu~\cite{sun2023emu} & LLaMA 13B & - & - & - & - & - \\

DreamLLM~\cite{dong2023dreamllm} & Vicuna 7B & - & - & - & - & 36.6 \\

Chameleon~\cite{team2024chameleon} & From Scratch 7B & - & - & - & 22.4 & 8.3 \\

Show-o~\cite{xie2024show} & Phi-1.5 1.3B & 1097.2 & - & - & 26.7 & - \\

VILA-U~\cite{wu2024vila} & LLaMA-2 7B & 1401.8 & - & 59.0 & - & 33.5 \\

Emu3~\cite{wang2024emu3} & From Scratch 7B & - & 58.5 & 68.2 & 31.6 & 37.2 \\

MetaMorph~\cite{tong2024metamorph} & LLaMA-3 8B & - & 75.2 & 71.8 & - & - \\

TokenFlow-XL~\cite{qu2025tokenflow} & Qwen-2.5 14B & 1551.1 & 76.8 & 72.6 & 43.2 & 48.2 \\

Transfusion~\cite{zhou2024transfusion} & From Scratch 7B & - & - & - & - & - \\

LMFusion~\cite{shi2024lmfusion} & LLaVA-Next 8B & 1603.7 & 72.1 & 72.5 & 41.7 & - \\

Janus~\cite{wu2025janus} & DeepSeek-LLM 1.5B & 1338.0 & 69.4 & 63.7 & 30.5 & 34.3 \\

JanusFlow~\cite{ma2025janusflow} & DeepSeek-LLM 1.5B & 1333.1 & 74.9 & 70.5 & 29.3 & 30.9 \\

Janus-Pro-1B~\cite{chen2025januspro} & DeepSeek-LLM 1.5B & 1444.0 & 75.5 & 68.3 & 36.3 & 39.8 \\

Janus-Pro-7B & DeepSeek-LLM 7B & 1567.1 & 79.2 & 72.1 & 41.0 & 50.0 \\

\textbf{Ours} & \textbf{Qwen2.5-VL 3B} & \textbf{1574.3} & \textbf{78.6} & \textbf{73.8} & \textbf{53.1} & \textbf{63.2} \\

\bottomrule

\end{tabular}

\end{table}

As a research project, our primary goal is to establish a simple, reproducible, and community-friendly baseline. The base model's competitive performance demonstrates its potential as a starting point for exploring unified RL strategies. Its lightweight design and reliance on open-source datasets ensure accessibility, encouraging further innovation in integrating LLMs and DMs for generative AI tasks.

\subsection{RL on text-to-image generation}
\label{subsec:rl_text_to_image}
\begin{figure}[h]
\centering
\includegraphics[width=0.95\textwidth]{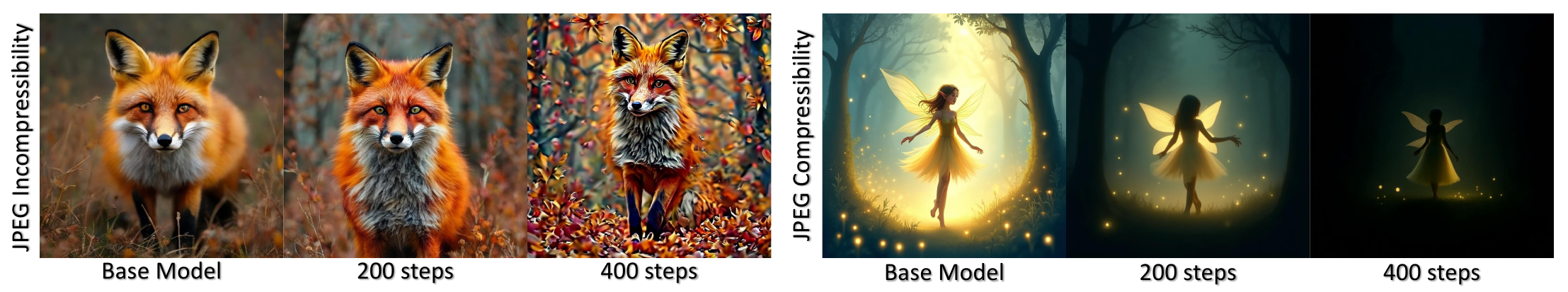}
\caption{Visual examples of image changes during RL training under JPEG compressibility and incompressibility rewards.}
\label{fig:jpeg_changes}
\end{figure}

To validate the effectiveness of UniRL-Zero for text-to-image generation, we initially employ two contrasting reward models: JPEG compressibility (higher rewards for smaller compressed file sizes) and JPEG incompressibility (higher rewards for larger compressed file sizes). This dual evaluation helps isolate implementation factors, ensuring the model can optimize rewards under opposing objectives. We observe that, during training, the storage size of JPEG-compressed generated images changes as expected: under JPEG compressibility, the file size decreases progressively, while under JPEG incompressibility, it increases. Visual changes in the generated images are illustrated in Figure~\ref{fig:jpeg_changes}.
To further assess our method, we conduct larger-scale, longer-duration training on the GenEval benchmark. We use a training set of 50,000 randomly generated GenEval prompts (ensuring no overlap with the test set) following Flow-GRPO. As shown in Table~\ref{tab:geneval}, Our experiments demonstrate effective improvements in GenEval score, confirming the robustness of our RL strategy for text-to-image generation. 

\begin{table}[h]
\centering
\caption{Performance comparison on GenEval dataset}\label{tab:geneval}
\resizebox{0.95\textwidth}{!}{
\begin{tabular}{l c c c c c c c}
\toprule
\textbf{Model/Configuration}  & \textbf{Overall} & \textbf{Single Object} & \textbf{Counting} & \textbf{Colors} & \textbf{Two Object} & \textbf{Position} & \textbf{Color Attribution} \\
\midrule
Base Model & 0.69 & 1.00 & 0.62 & 0.87 & 0.85 & 0.32 & 0.58 \\
T2I-RL &0.80 & 1.00 & 0.80 & 0.91 & 0.96 & 0.35 & 0.79 \\
CoT-enhanced T2I-RL &  0.85 & 1.00 & 0.85 & 0.93 & 0.96 & 0.50 & 0.80 \\
\bottomrule
\end{tabular}}
\end{table}
\subsection{RL on CoT-enhanced text-to-image generation}
\label{subsec:rl_cot_text_to_image}
\begin{figure}[h]
\centering
\begin{subfigure}[t]{0.48\textwidth}
    \centering
    \includegraphics[width=\textwidth]{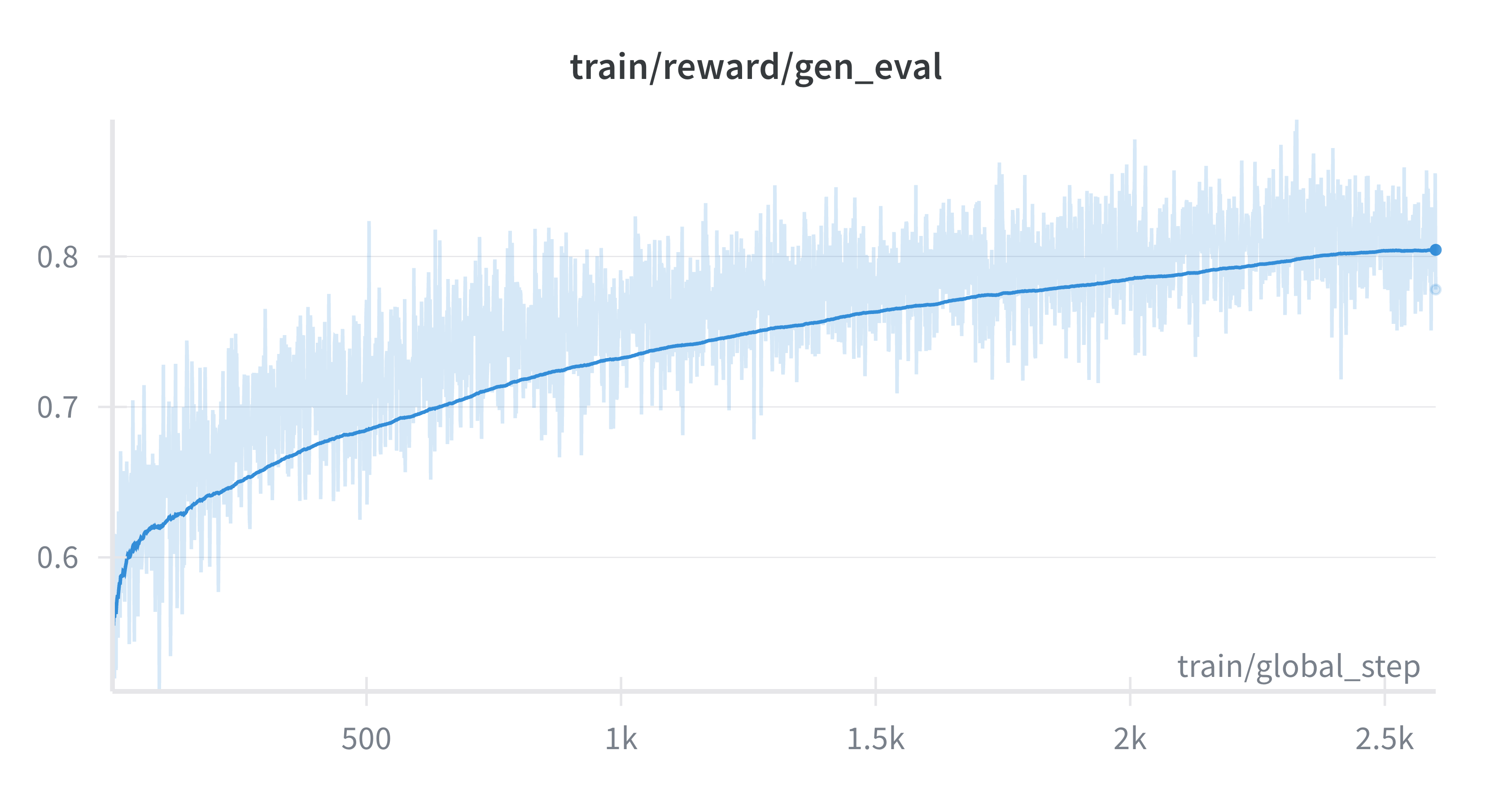} 
    \caption{Rewards of GenEval during training.}
    \label{fig:geneval_curves}
\end{subfigure}
\hfill
\begin{subfigure}[t]{0.48\textwidth}
    \centering
    \includegraphics[width=\textwidth]{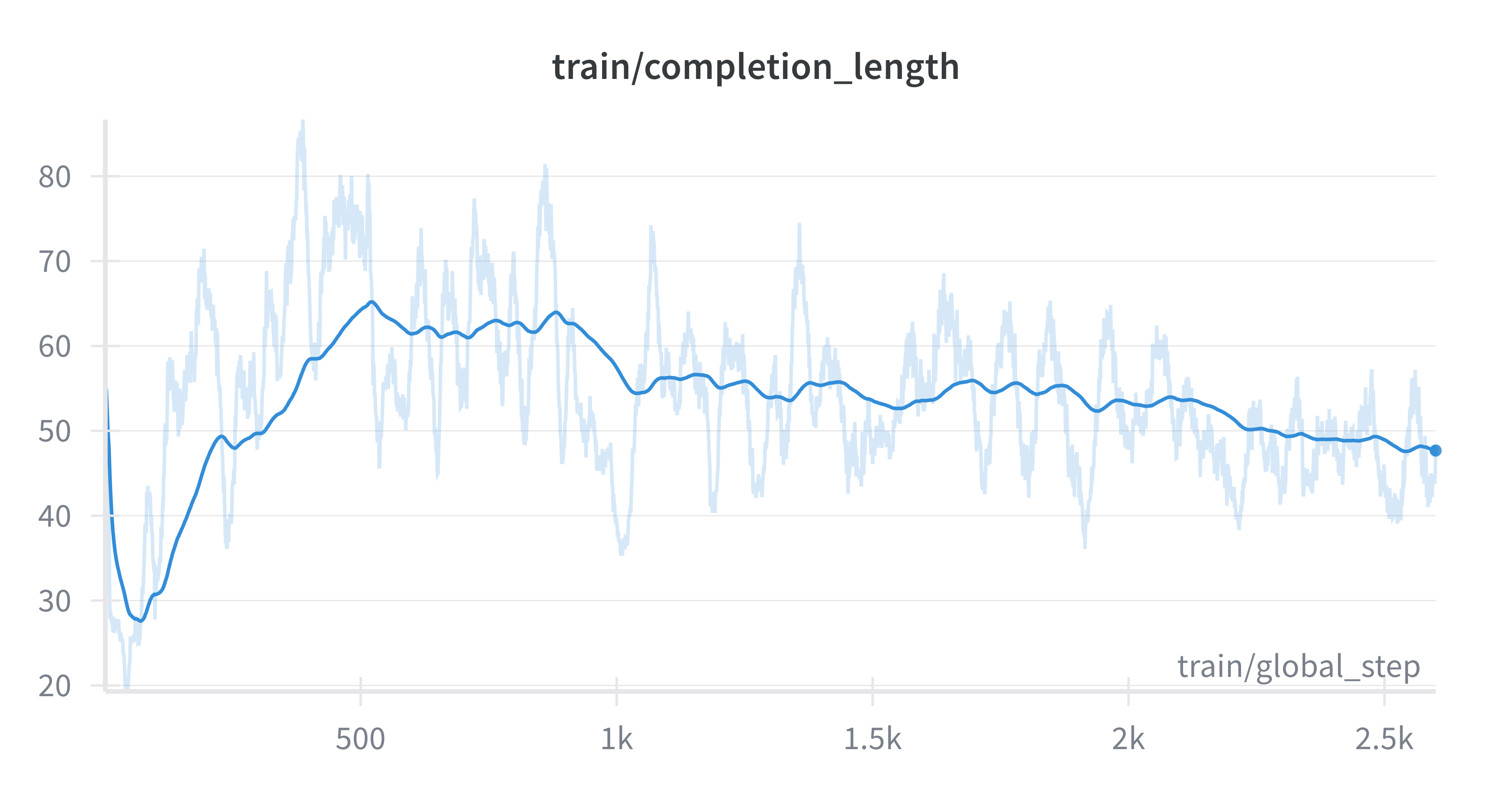} 
    \caption{Reasoning lengths throughout training.}
    \label{fig:reasoning_length_curves}
\end{subfigure}
\caption{Training curves of the CoT-enhanced text-to-image generation.}\label{fig:cot_curves}
\end{figure}

\begin{figure}[h]
    \centering
    \includegraphics[width=\textwidth]{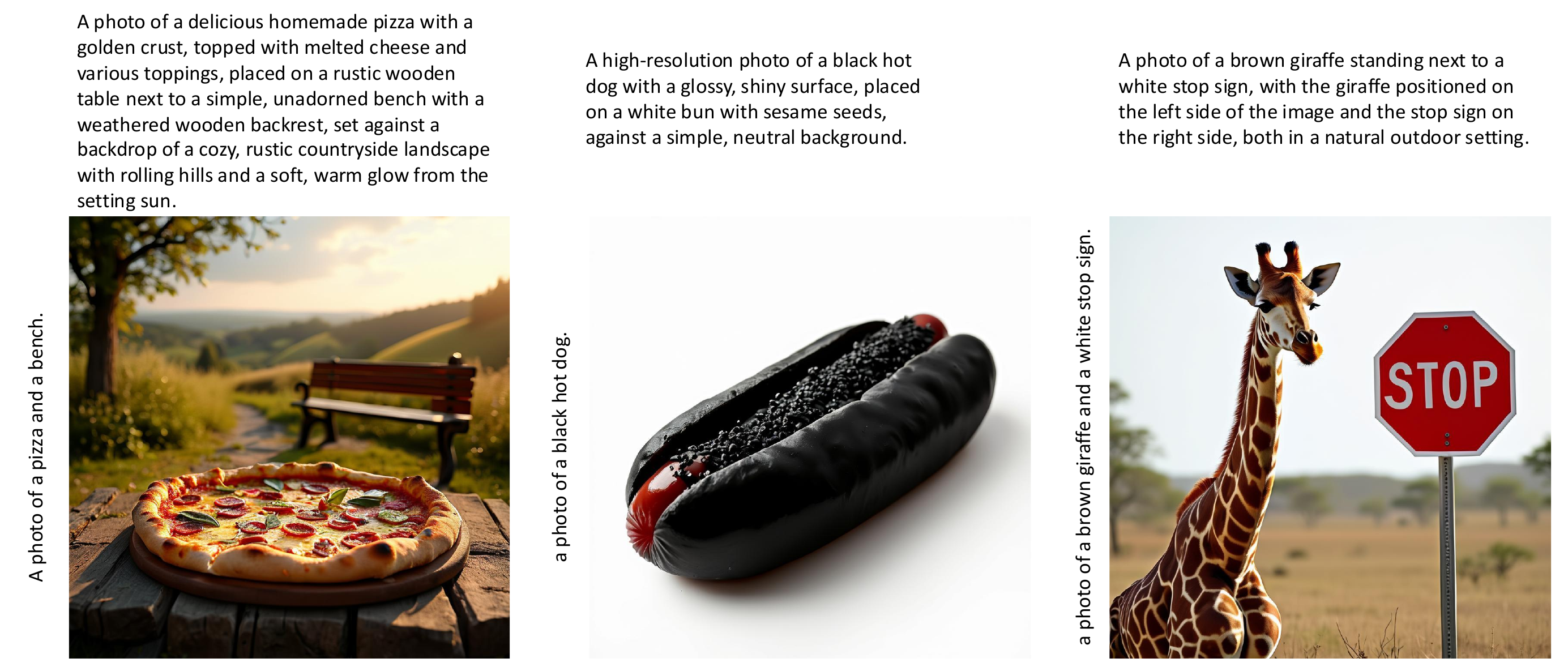}
    \caption{Generation examples of the CoT-enhanced text-to-image generation.  Original prompts are on the left of images and improved prompts are above images.}
    \label{fig:cot-example}
\end{figure}

In real-world scenarios, users often provide brief or vague prompts, yet high-quality image generation typically requires precise and detailed prompts. CoT-enhanced text-to-image generation leverages the reasoning capabilities of the multimodal language model to generate refined prompts for improved image synthesis. Notably, our unified model was pretrained solely on text-to-image and instructional image editing datasets, lacking explicit support for reasoning-enhanced generation.
However, we find that a small-scale cold start fine-tuning of the DM (with the LM kept frozen) using a limited dataset of reasoning-enhanced image-text pairs enables the model to perform precise reasoning-driven image generation. We provide more details in Supplementary Section~\ref{sec:cold-start}. Through RL, we further amplify the benefits of this reasoning capability. For training, we use 50,000 randomly generated GenEval prompts. The LM first enhances the original prompt with reasoning, followed by context extraction for image generation via the DM. Our results show: (1) improved performance on GenEval metrics and (2) dynamic adaptation in the length and complexity of reasoning outputs, as illustrated in Figure~\ref{fig:cot_curves}.  The evaluation results are shown in Table~\ref{tab:geneval}. We show some generation examples in Figure~\ref{fig:cot-example}.


\begin{figure}[t]
\centering
\includegraphics[width=\textwidth]{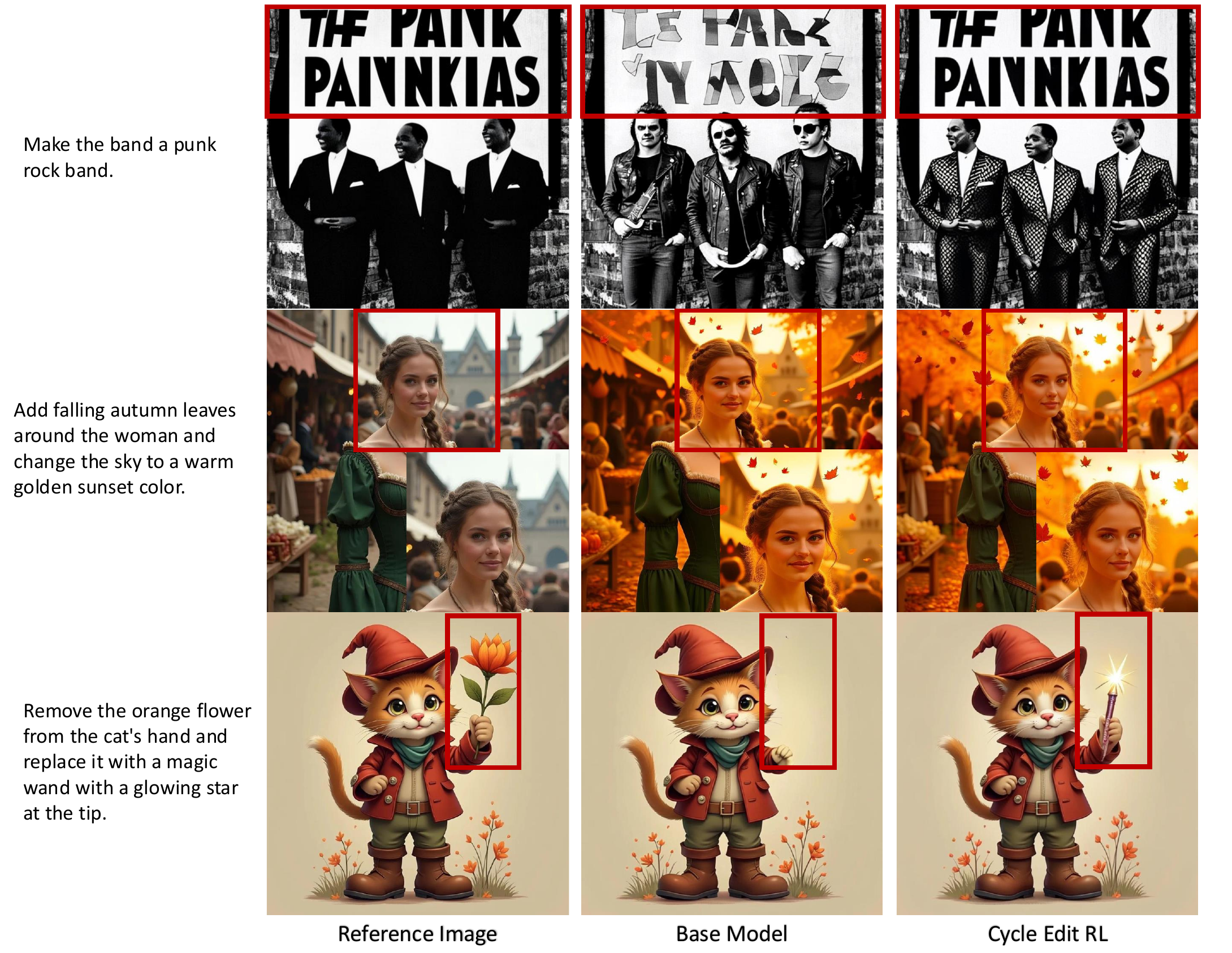}
\caption{Visual examples showcasing the effectiveness of Cycle Edit RL, highlighting enhanced similarity to the reference image and improved instrunction following.}
\label{fig:cycle_edit_examples}
\end{figure}

\subsection{RL on instructional image editing}
\label{subsec:rl_image_editing}

The primary goal of instructional image editing is to produce an edited image that closely aligns with the user’s editing instructions while preserving the structural and visual similarity to the reference image. This dual objective ensures that edits are both instruction-compliant and contextually consistent with the original content. To achieve this, we propose \textbf{Cycle Edit RL}, a reinforcement learning (RL) approach inspired by CycleGAN's cycle consistency~\cite{zhu2017unpaired}.

\textbf{Cycle Edit RL} processes a reference image \(\mathbf{x}_{\text{ref}}\) with an editing instruction \(I_{\text{edit}}\) (e.g., ``Add a forest background'') to produce an edited image \(\mathbf{x}_{\text{edit}}\). A reverse instruction \(I_{\text{reverse}}\) (e.g., ``Remove the forest background'') is then applied to \(\mathbf{x}_{\text{edit}}\) to generate a cycled image \(\mathbf{x}_{\text{cycle}}\), aiming to closely match \(\mathbf{x}_{\text{ref}}\).

Formally, the Cycle Edit RL process consists of the following steps:
\begin{enumerate}
    \item \textbf{Forward Edit}: The unified model processes the input \(\mathcal{Q} = (I_{\text{edit}}, \mathbf{x}_{\text{ref}})\) to generate the edited image \(\mathbf{x}_{\text{edit}}\).
    \item \textbf{Reverse Edit}: The model processes \(\mathcal{Q}' = (I_{\text{reverse}}, \mathbf{x}_{\text{edit}})\) to produce the cycled image \(\mathbf{x}_{\text{cycle}}\), using the reverse instruction \(I_{\text{reverse}}\).
    \item \textbf{Cycle Consistency Reward}: We use CLIP to measure the similarity between the reference and cycled images:
    \begin{equation}
    R_{\text{cycle}} = \text{CLIP}(\mathbf{x}_{\text{ref}}, \mathbf{x}_{\text{cycle}}).
    \end{equation}
    \item \textbf{Total Reward}: The trajectory reward combines multiple components:
    \begin{equation}
    R(\tau) = \lambda_1 R_{\text{edit}}(\mathbf{x}_{\text{edit}}, I_{\text{edit}}) + \lambda_2 R_{\text{cycle}}(\mathbf{x}_{\text{ref}}, \mathbf{x}_{\text{cycle}}) + \lambda_3 R_{\text{quality}}(\mathbf{x}_{\text{edit}}),
    \end{equation}
    where \(R_{\text{edit}}\) evaluates instruction alignment using CLIP text-image direction similarity (inspired by InstructPix2Pix~\cite{brooks2023instructpix2pix}), \(R_{\text{quality}}\) assesses image quality, and \(\lambda_1, \lambda_2, \lambda_3\) are pre-defined hyperparameters.
\end{enumerate}

To construct the RL dataset, we select high-quality image-text pairs, including captions and reference images. We prompt Claude to generate creative editing instructions and infer corresponding reverse instructions, leveraging Claude’s reasoning capabilities without generating actual images. The resulting dataset comprises 200 curated training samples.

\begin{figure}[h]
    \centering
    \includegraphics[width=\textwidth]{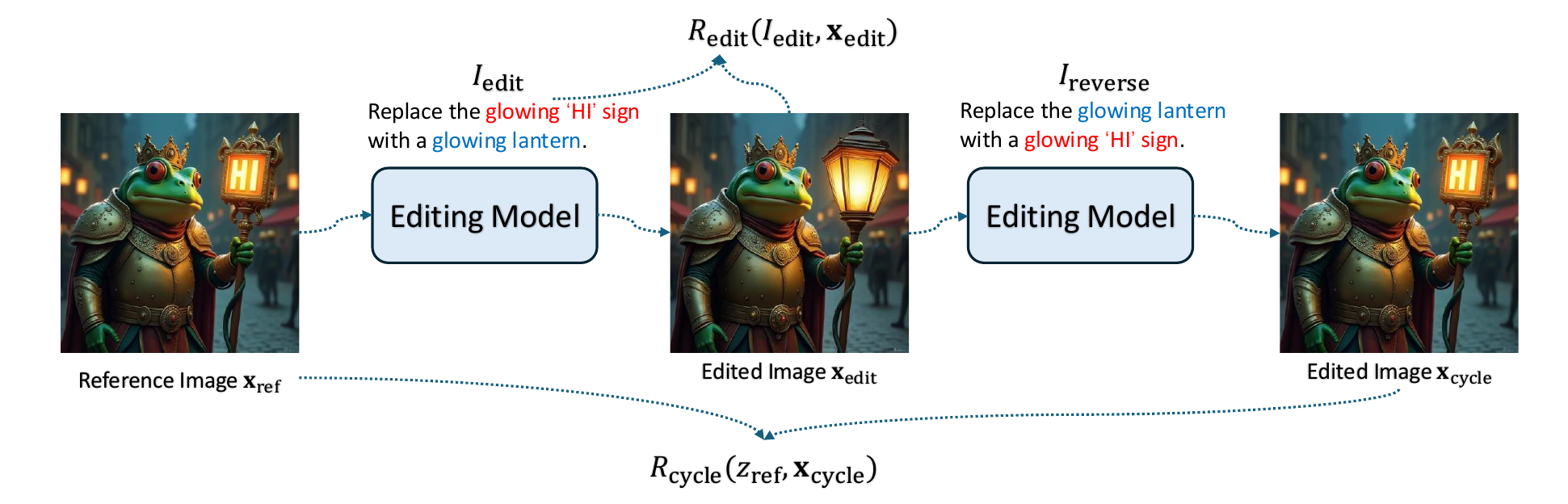}
    \caption{Illustration of the training pipeline for Cycle Edit RL.}
    \label{fig:cycle_edit}
\end{figure}

\begin{figure}[h]
    \centering
    \begin{subfigure}{0.45\linewidth}
        \centering
        \includegraphics[width=\linewidth]{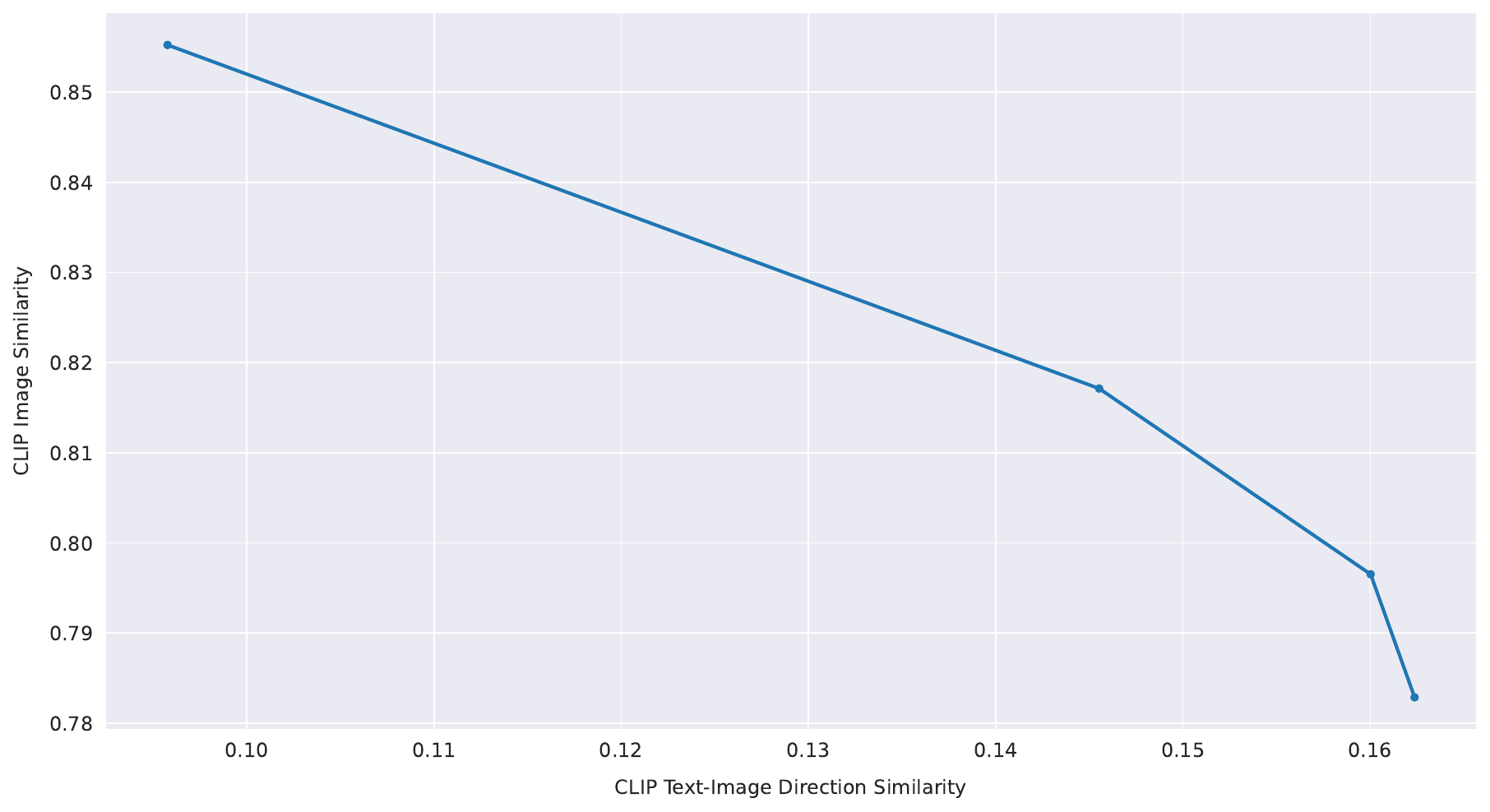}
        \caption{Base model}
        \label{fig:sub1}
    \end{subfigure}
    \hfill
    \begin{subfigure}{0.45\linewidth}
        \centering
        \includegraphics[width=\linewidth]{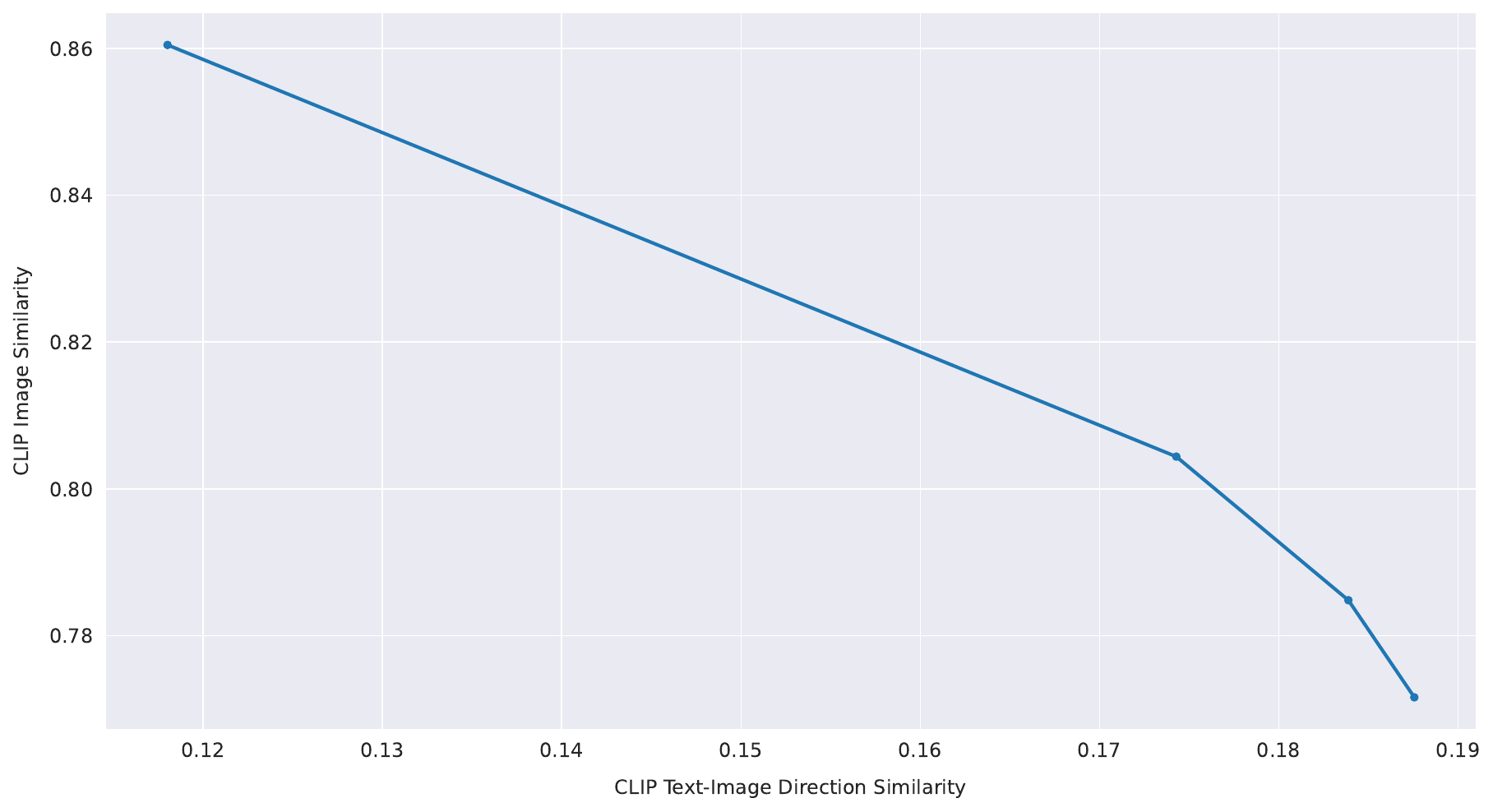}
        \caption{Cycle Edit RL}
        \label{fig:sub2}
    \end{subfigure}
    \caption{Trade-off between consistency with the input image (Y-axis) and consistency with the edit instruction (X-axis), with text guidance varied at 1, 3, 5, and 7.}
    \label{fig:cycle_edit_metrics}
\end{figure}

The reward function uses CLIP-based metrics: \(R_{\text{edit}}\) evaluates instruction alignment, while \(R_{\text{cycle}}\) measures image similarity between \(\mathbf{x}_{\text{ref}}\) and \(\mathbf{x}_{\text{cycle}}\). The overall pipeline is illustrated in Figure~\ref{fig:cycle_edit}. As shown in Figure~\ref{fig:cycle_edit_metrics}, Cycle Edit RL significantly improves the model’s ability to follow editing instructions while preserving similarity to the reference image. Additional visual examples in Figure~\ref{fig:cycle_edit_examples} demonstrate enhanced structural and detail retention in the edited images.

\begin{figure}[h]
\centering
\includegraphics[width=0.4\textwidth]{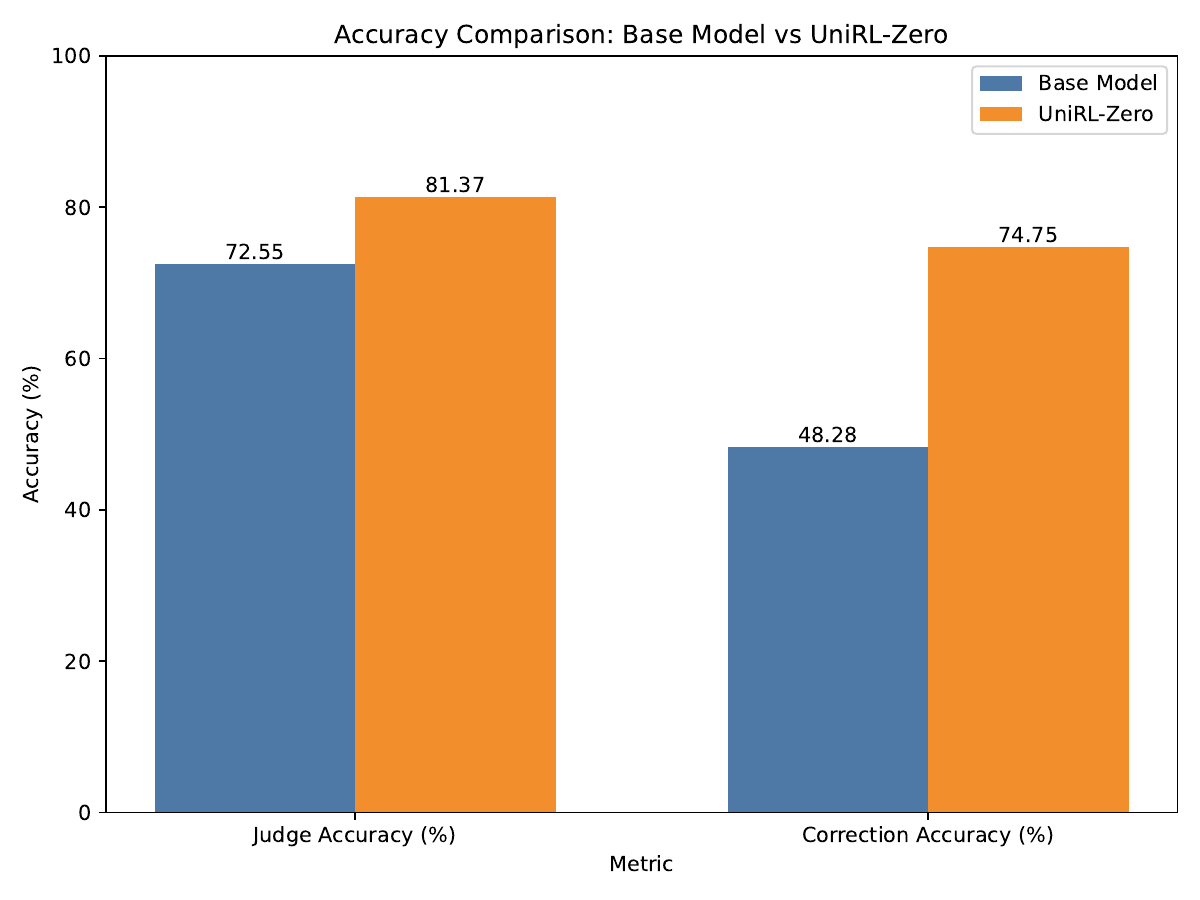}
\caption{The effectiveness of the RL training on improving the model’s reflection accuracy and the correction rate of erroneous image.}
\label{fig:reflection}
\end{figure}

\begin{figure}[h]
\centering
\includegraphics[width=0.9\textwidth]{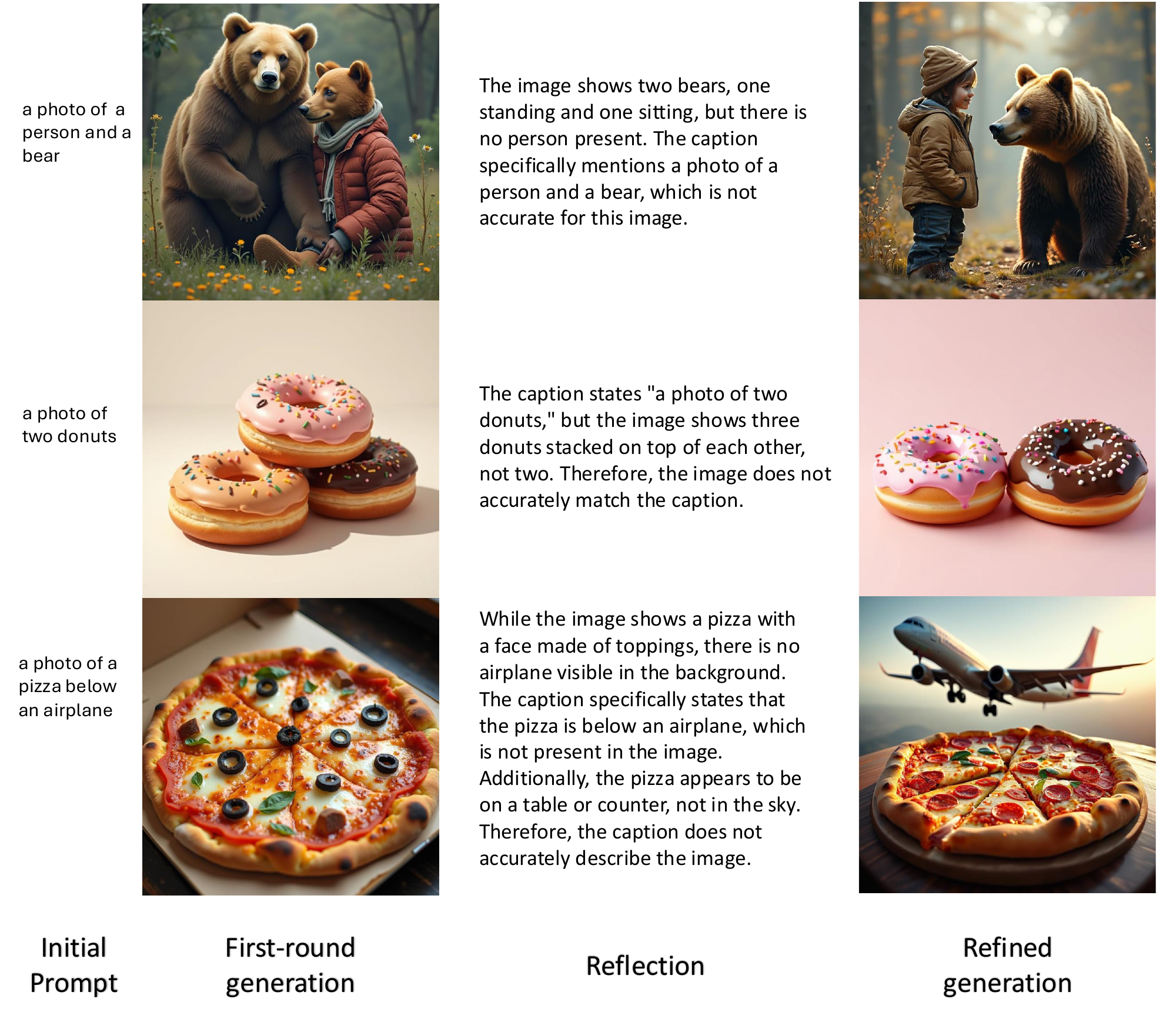}
\caption{Visual examples of image generation reflection, demonstrating improved error identification and correction.}
\label{fig:reflection_examples}
\end{figure}

\subsection{RL on image generation reflection}
\label{subsec:rl_image_reflection}
In unified models, the DM generates images by leveraging the context vectors extracted from the LM. Due to the inherent randomness and instability of the generation process, DM may not always accurately fulfill all prompt requirements in a single attempt. A unified model should ideally (1) reflect on inaccuracies in generated images and (2) refine them accordingly. Leveraging the LM’s multimodal understanding and analysis capabilities, we use RL to enhance: (1) the model’s accuracy in identifying generation errors and (2) its ability to correct flawed images.
To construct the RL dataset, we select GenEval prompts and generate multiple images per prompt using Flux. We then apply GenEval’s detection logic to identify correct and incorrect images. To mitigate potential errors in GenEval’s detection tools (\eg, false negatives), we employ Claude for additional validation, ensuring high-quality pairs. This results in a dataset where each prompt has both correct and incorrect image outputs.
Prior to RL training, we fine-tune the DM using 10k reflection-augmented data points as the cold start. We find that computing autoregressive (AR) loss for the LM leads to rapid degradation of its understanding capabilities, so we avoid this.
For RL training, we use approximately 1,500 data points. For each generated image, the LM evaluates whether it matches the prompt, responding with a Yes'' or No'' answer. We assign a reward of 0-1 based on the correctness of this response and an additional 0-1 reward for proper response formatting. For corrected images, we use GenEval to score the alignment with the prompt. As shown in Figure~\ref{fig:reflection}, RL training effectively improves the model’s reflection accuracy (Judge Accuracy) and the correction rate of erroneous images (Correction Accuracy). Visual examples in Figure~\ref{fig:reflection_examples} illustrate this process.

\subsection{Conclusion and limitations}
This paper introduced UniRL-Zero, a unified reinforcement learning framework on unified understanding and generation models. We defined six core scenarios that cover understanding, generation, and their beneficial interactions. We trained a simple base unified model with competitive results on both multimodal understanding and generation. On top of this, our experiments validate the feasibility and effectiveness of RL training on the unified model, demonstrating improvements in instruction adherence, compositional accuracy, and editing consistency. These results establish UniRL-Zero as a robust foundation for advancing unified RL frameworks, particularly in complex generative tasks requiring tight LM-DM synergy.

\paragraph{Limitations.}
Despite the gains, several limitations remain:
\begin{itemize}
\item \textbf{Reward bias.} Rewards such as CLIP-based alignment and GenEval metrics do not cover all aspects of quality (scene geometry, long-range coherence, fine-grained attributes). Our work, as a research project, mainly aims to validate the effectiveness of RL training rather than to design comprehensive reward models. More diverse and fine-grained reward functions might be required for broader applicability.  

\item \textbf{Experimental scale.} Due to limited computational resources, our experiments were conducted at a relatively modest scale in terms of data volume, model size, and training duration. While the base unified model already shows competitive performance, its capacity is still insufficient compared to large-scale proprietary systems. As a result, the improvements demonstrated here may underestimate the potential of UniRL-Zero under large-scale training. Scaling up data coverage, diffusion expert size, and RL training steps will likely yield stronger results.  
\end{itemize}

{
    \small
    \bibliographystyle{ieeenat_fullname}
    \bibliography{main}
}

\clearpage

\setcounter{section}{0}  
\renewcommand{\thesection}{\Alph{section}}  
\onecolumn

\section*{\Large Supplementary}

\section{Training a Unified Base Model}\label{sec:basemodel}

\subsection{Model architecture}

The unified model integrates a multimodal language model (LM) and a diffusion model (DM) through specialized connectors and query tokens:

\begin{itemize}
    \item \textbf{LLM}: We adopt the pretrained Qwen2.5-VL-Instruct~\cite{bai2025qwen2} as the frozen multimodal backbone, preserving its robust understanding and reasoning capabilities across modalities.
    \item \textbf{DM}: A linear transformer based on SANA-1.6B~\cite{xie2024sana} serves as the diffusion expert for generation. High-resolution images are encoded into low-resolution latents using a pretrained DC-VAE~\cite{chen2025deepcompressionautoencoderefficient} for training efficiency. For image editing, the DM's input layer is expanded to incorporate reference image latents, enhancing editing quality. These additional channels are set to zero for text-to-image tasks.
    \item \textbf{Query tokens}: Two sets of lightweight query vectors are employed—one for text-to-image generation and another for image editing—to accommodate distinct instruction and caption styles, adding minimal parameters.
    \item \textbf{Connectors}: A bidirectional attention transformer processes LM-extracted query features, feeding them to the DM via cross-attention, enabling seamless interaction between the LM and DM.
\end{itemize}

\subsection{Training data}

Leveraging the frozen LM, we do not need to collect vast amounts of high-quality text reasoning and vision understanding data to maintain the LM’s reasoning and understanding capabilities. We focus on curating datasets for image generation and editing, using open-source resources to ensure reproducibility:

\begin{itemize}
    \item \textbf{Text-to-image}: We utilize the \href{https://huggingface.co/datasets/jackyhate/text-to-image-2M}{text-to-image-2M} and \href{https://huggingface.co/datasets/lehduong/flux_generated}{flux\_generated} datasets, comprising ~3.5M image-text pairs. These are filtered using PickScore, retaining the top 50\% for pretraining. To enhance compositional ability, we generate ~50K images using Flux, guided by GenEval prompts, and filter them to meet GenEval's detection criteria.
    \item \textbf{Image editing}: We employ ~1.2M edited image pairs from the OmniEdit~\cite{wei2025omnieditbuildingimageediting} dataset. Additionally, we select the top 100K images from \texttt{flux\_generated} based on PickScore~\cite{NEURIPS2023_73aacd8b} and use Claude to generate creative editing instructions. Edited results are produced using the Flux-Kontext~\cite{labs2025flux1kontextflowmatching} model and combined with OmniEdit for training.
\end{itemize}

The combined dataset includes 1.8M image-text pairs and 1.3M instructional editing pairs, trained for 10 epochs with uniform sampling.

\subsection{Training details}

The DM follows a linear diffusion process: \(\mathbf{x}_t = (1-t)\mathbf{x}_0 + t\boldsymbol{\epsilon}\), where \(\mathbf{x}_0\) is the training data and \(\boldsymbol{\epsilon} \sim \mathcal{N}(\boldsymbol{0}, \mathbf{I})\). We use v-prediction, defined as \(\boldsymbol{v} = \boldsymbol{\epsilon} - \mathbf{x}_0\), and sample timesteps \(t\) via \(t = \text{torch.sigmoid}(\text{torch.normal}(\text{mean}=0, \text{std}=1, \text{size}=(\text{bsz},)))\), following Stable Diffusion 3, with unit weights.

For image editing, reference image latents are noised as \(\hat{\boldsymbol{z}}_{\text{ref}} = 0.8\boldsymbol{z}_{\text{ref}} + 0.2\boldsymbol{\epsilon}'\), where \(\boldsymbol{\epsilon}' \sim \mathcal{N}(0, \mathbf{I})\). For text-to-image tasks, text captions are masked with 10\% probability to support classifier-free guidance. For image editing, we mask text captions (5\% probability), reference image latents (5\% probability), or both (5\% probability).

\subsection{Cold Start}\label{sec:cold-start}

The pretraining datasets primarily consist of general image-text pairs for text-to-image generation and instructional image editing. To enhance performance in complex scenarios, such as reasoning-heavy or error-prone generation tasks, we fine-tune the DM using targeted datasets, keeping the LM frozen to preserve its capabilities:

\begin{itemize}
    \item \textbf{Chain-of-Thought Enhanced Generation (Scenario 5)}: To improve text-to-image generation under complex reasoning, we select 10K image-text pairs with short captions from the pretraining dataset. Using Claude, we generate detailed captions that incorporate chain-of-thought reasoning to better describe image content. The DM is fine-tuned on these 10K pairs for 10K steps, optimizing its ability to generate images from nuanced text inputs.
    \item \textbf{Image Generation Reflection~(Scenario 6)}: To address errors in compositional image generation, we use GenEval prompts to generate multiple images per prompt with Flux~\cite{flux2024}. We select pairs containing both correct and incorrect outputs. Claude analyzes the incorrect images, identifying errors and proposing improvements by referencing correct images. This results in approximately 10K prompt-image-reasoning-refined image pairs, which are used to fine-tune the DM, enhancing its robustness in complex generation tasks.
\end{itemize}

Training loss is computed solely for image generation, with gradients applied only to the DM. The LM's parameters remain unchanged, as supervised fine-tuning of the LM was found to degrade its understanding and reasoning capabilities. This approach ensures the DM adapts effectively to complex LM-generated contexts without compromising the LM's performance.

\section{SDE sampling for flow matching models}\label{sec:sdeproof}

The sampling process of the diffusion SDE is equivalent to first adding nose and then denoise using Euler solver, here we provide an intuitive proof showcasing our implementation.

Given a $\mathbf x_t$ which is denoised from $\mathbf x_1 \sim \mathcal N(\boldsymbol 0, \mathbf I)$, we firstly add noise to $\mathbf x_{t + Dt}$. Following the derivation of variational diffusion model (A.1 of supplementary), we have 
\begin{equation}
\begin{split}
q(\mathbf x_{t+Dt}|\mathbf x_t) & = \mathcal N(\alpha_{(t + Dt) \mid t} \mathbf x_t, \sigma_{(t+Dt)\mid t}^2 \mathbf I )\\
\alpha_{(t+Dt)\mid t} & = \frac{\alpha_{t+Dt}}{\alpha_t} = \frac{1-t -Dt}{1-t} = 1-\frac{Dt}{1-t} \\
\sigma_{(t+Dt) \mid t}^2 & = \sigma_{t+Dt}^2 - (\frac{\alpha_{t+Dt}}{\alpha_t})^2 \sigma_{t}^2 = t^2 + 2Dt \cdot t + Dt^2 - (1-2\frac{Dt}{1-t} + (\frac{Dt}{1-t})^2) t^2
\end{split}
\end{equation}
 Assuming $Dt\rightarrow 0$, we can ignore second-order infinitesimal $Dt^2$ and have 
\begin{equation}
\sigma_{(t+Dt) \mid t}^2 = t^2 + 2Dt \cdot t - t^2 +2\frac{Dt}{1-t} t^2 = \frac{2 Dt \cdot t}{1-t}
\end{equation}
Thus, considering the re-parameterization trick,  we have
\begin{equation}
\mathbf x_{t+Dt} = (1-\frac{Dt}{1-t})\mathbf x_t + \sqrt{\frac{2 Dt \cdot t}{1-t}} \boldsymbol \epsilon, \quad \boldsymbol \epsilon \sim \mathcal N(\boldsymbol 0,\mathbf I)\, .
\end{equation}
We then denoise from $\mathbf x_{t+Dt}$ to $\mathbf x_{t+dt}$ following the velocity, where $Dt > 0$ and $dt < 0$, 
\begin{equation}
\mathbf x_{t+dt} = \mathbf x_{t+Dt} + \int_{t+Dt}^{t+dt} \boldsymbol v_s \mathrm ds \approx \mathbf x_{t+Dt} + \boldsymbol v_t (dt - Dt)
\end{equation}
The above approximation can be obtained by using first-order taylor expansion by ignoring the high order error terms. We provide a simple proof in the following

\textbf{Proof using integral decomposition and Taylor expansion:}

Given $t+Dt < t < t+dt$, decompose the integral:
\begin{equation}
\int_{t+Dt}^{t+dt} \boldsymbol{v}_s \, ds = \int_{t+Dt}^{t} \boldsymbol{v}_s \, ds + \int_{t}^{t+dt} \boldsymbol{v}_s \, ds
\end{equation}

\textbf{For the first integral:}
Let $F_1(u) = \int_{u}^{t} \boldsymbol{v}_s \, ds$, so $F_1'(u) = -\boldsymbol{v}_u$.

Taylor expansion around $u = t$:
\begin{equation}
\int_{t+Dt}^{t} \boldsymbol{v}_s \, ds = F_1(t+Dt) \approx F_1(t) + F_1'(t) \cdot Dt = 0 + (-\boldsymbol{v}_t) \cdot Dt = -\boldsymbol{v}_t Dt
\end{equation}

\textbf{For the second integral:}
Let $F_2(u) = \int_{t}^{u} \boldsymbol{v}_s \, ds$, so $F_2'(u) = \boldsymbol{v}_u$.

Taylor expansion around $u = t$:
\begin{equation}
\int_{t}^{t+dt} \boldsymbol{v}_s \, ds = F_2(t+dt) \approx F_2(t) + F_2'(t) \cdot dt = 0 + \boldsymbol{v}_t \cdot dt = \boldsymbol{v}_t dt
\end{equation}

\textbf{Combined result:}
\begin{equation}
\int_{t+Dt}^{t+dt} \boldsymbol{v}_s \, ds = -\boldsymbol{v}_t Dt + \boldsymbol{v}_t dt = \boldsymbol{v}_t(dt - Dt)
\end{equation}
Therefore, we have our SDE sampling step is given by 
\begin{equation}
\begin{split}
\mathbf x_{t+dt} & = \mathbf x_{t+Dt} + \int_{t+Dt}^{t+dt} \boldsymbol v_s \mathrm ds \\ & \approx \mathbf x_{t+Dt} + \boldsymbol v_t (dt - Dt) \\
& = (1-\frac{Dt}{1-t})\mathbf x_t + \sqrt{\frac{2 Dt \cdot t}{1-t}} \boldsymbol \epsilon +  \boldsymbol v_t (dt - Dt)
\end{split}
\end{equation}
Again, through the re-parameterization trick, we can have 

\begin{equation}
    \mathbf x_{t+dt} \sim \mathcal N((1-\frac{Dt}{1-t})\mathbf x_t +  \boldsymbol v_t (dt - Dt), \frac{2 Dt \cdot t}{1-t}) \, .
\end{equation}

\subsection{Equivalent but simpler implementation of SDE sampling to Flow-GRPO}
\begin{lstlisting}[caption={SDE sampling implementation comparison}]
def sde_step_flowgrpo(
    model_output: torch.FloatTensor,
    sigma: torch.FloatTensor,
    dt: torch.FloatTensor,
    sample: torch.FloatTensor,
):
    eta = 1.0
    std_dev_t = torch.sqrt(sigma / (1 - torch.where(sigma == 1, 0.9931, sigma))) * eta
    prev_sample_mean = sample*(1+std_dev_t**2/(2*sigma)*dt) + \
        model_output*(1+std_dev_t**2*(1-sigma)/(2*sigma))*dt
    variance_noise = randn_tensor(
        model_output.shape,
        device=model_output.device,
        dtype=model_output.dtype,
    )
    prev_sample = prev_sample_mean+ std_dev_t * torch.sqrt(-1*dt)* variance_noise
    log_prob = (
        -((prev_sample.detach() - prev_sample_mean) ** 2) / (2 * ((std_dev_t * torch.sqrt(-1*dt))**2))
        - torch.log(std_dev_t * torch.sqrt(-1*dt))
        - torch.log(torch.sqrt(2 * torch.as_tensor(math.pi)))
    )
    log_prob = log_prob.mean(dim=tuple(range(1, log_prob.ndim)))
    return prev_sample, log_prob, prev_sample_mean, std_dev_t * torch.sqrt(-1*dt)

def sde_step_ours(
    model_output: torch.FloatTensor,
    sigma: torch.FloatTensor,
    dt: torch.FloatTensor,
    sample: torch.FloatTensor,
):
    eta_squared_div_2 = 0.5
    Dt = - dt * eta_squared_div_2
    prev_sample_mean = sample * (1 - Dt / (1 - torch.where(sigma == 1, 0.9931, sigma))) + model_output * (dt - Dt)
    std_dev_t = torch.sqrt(2 * Dt * (sigma / (1 - torch.where(sigma == 1, 0.9931, sigma))))
    variance_noise = randn_tensor(
        model_output.shape,
        device=model_output.device,
        dtype=model_output.dtype,
    )
    prev_sample = prev_sample_mean +  std_dev_t * variance_noise
    log_prob = (
        -((prev_sample.detach() - prev_sample_mean) ** 2) / (2 * (std_dev_t**2))
        - torch.log(std_dev_t)
        - torch.log(torch.sqrt(2 * torch.as_tensor(math.pi)))
    )
    log_prob = log_prob.mean(dim=tuple(range(1, log_prob.ndim)))
    return prev_sample, log_prob, prev_sample_mean, std_dev_t

def main():
    test_basic_cases() # yes, equivalent
    test_extreme_cases() # yes, equivalent
\end{lstlisting}

\end{document}